\documentclass[10pt,twocolumn,letterpaper]{article}

\usepackage{iccv}
\usepackage{times}
\usepackage{epsfig}
\usepackage{graphicx}
\usepackage{amsmath}
\usepackage{amssymb}

\usepackage{comment}
\usepackage{multirow}

\usepackage[pagebackref=true,breaklinks=true,letterpaper=true,colorlinks,bookmarks=false]{hyperref}

\iccvfinalcopy % *** Uncomment this line for the final submission

\def\iccvPaperID{****} % *** Enter the ICCV Paper ID here

\ificcvfinal\fi

\begin{document}

%%%%%%%%% TITLE
\title{CutDepth:\\Edge-aware Data Augmentation in Depth Estimation}

\author{Yasunori Ishii\\
Panasonic\\
1006 Kadoma, Kadoma City, Osaka, Japan\\
{\tt\small ishii.yasunori@jp.panasonic.com}
\and
Takayoshi Yamashita\\
Chubu University\\
1200 Matsumotocho, Kasugai, Aichi, Japan \\
{\tt\small yamashita@isc.chubu.ac.jp}
}

\maketitle
\ificcvfinal\thispagestyle{empty}\fi

%%%%%%%%% ABSTRACT
\begin{abstract}
It is difficult to collect data on a large scale in a monocular depth estimation because the task requires the simultaneous acquisition of RGB images and depths.
Data augmentation is thus important to this task.
However, there has been little research on data augmentation for tasks such as monocular depth estimation, where the transformation is performed pixel by pixel.
In this paper, we propose a data augmentation method, called CutDepth.
In CutDepth, part of the depth is pasted onto an input image during training.
The method extends variations data without destroying edge features.
Experiments objectively and subjectively show that the proposed method outperforms conventional methods of data augmentation.
The estimation accuracy is improved with CutDepth even though there are few training data at long distances.
\end{abstract}

%%%%%%%%% BODY TEXT
%%%%%%%%%%%%%%%%%introduction
\begin{figure}[t]
\centering
\scalebox{0.9}{ %ƒRƒR
\begin{tabular}{ccc}
  \includegraphics[clip,width=0.28\linewidth,bb=0 0 320 240]{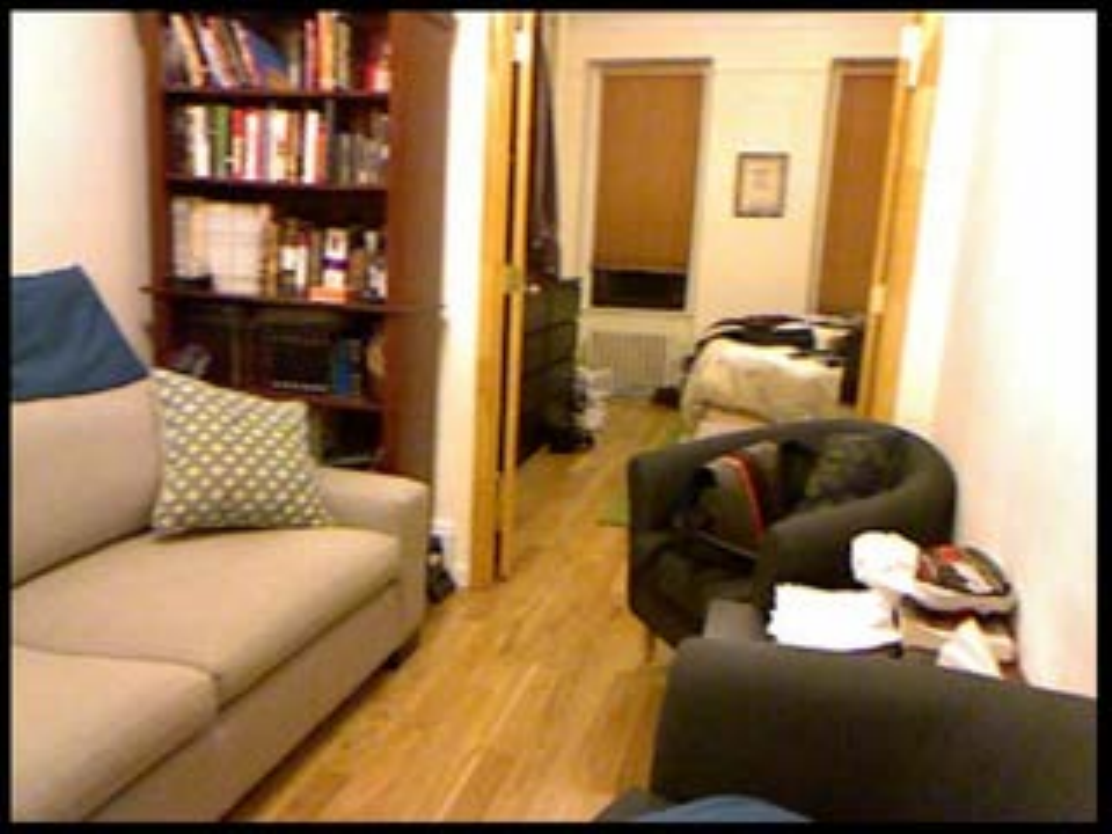} &
  \includegraphics[clip,width=0.28\linewidth,bb=0 0 320 240]{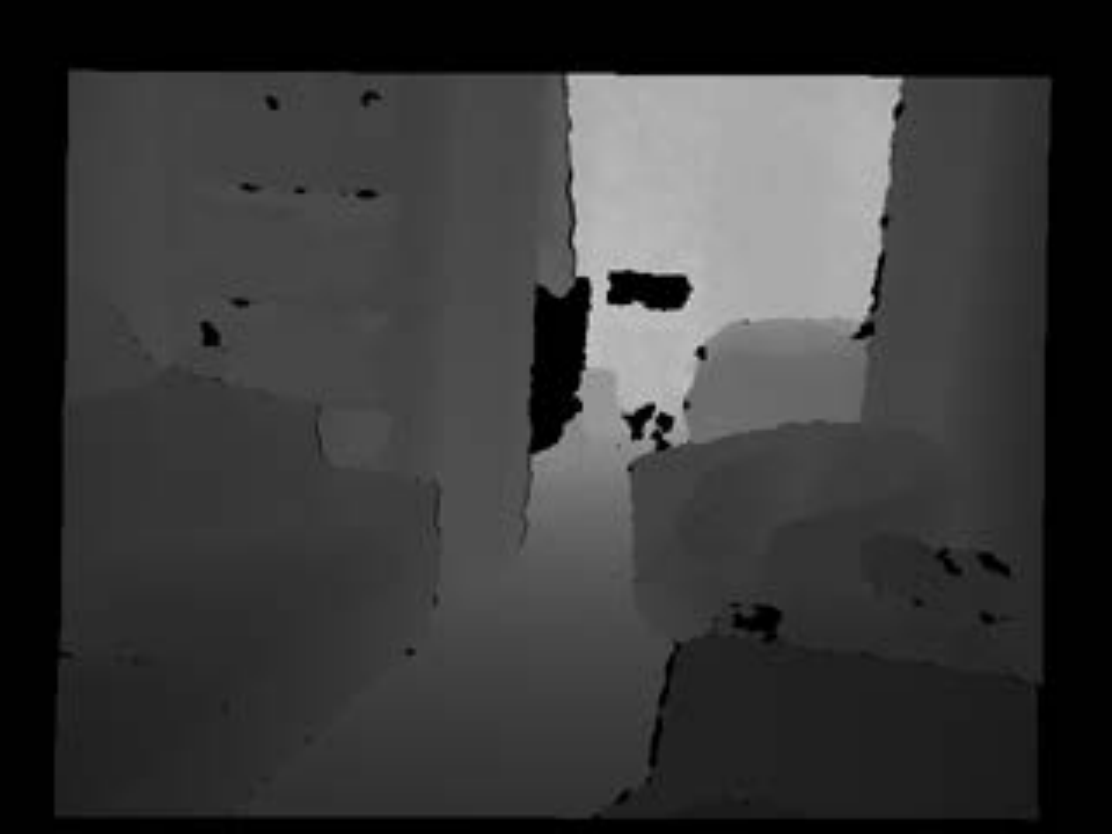} &
  \includegraphics[clip,width=0.28\linewidth,bb=0 0 320 240]{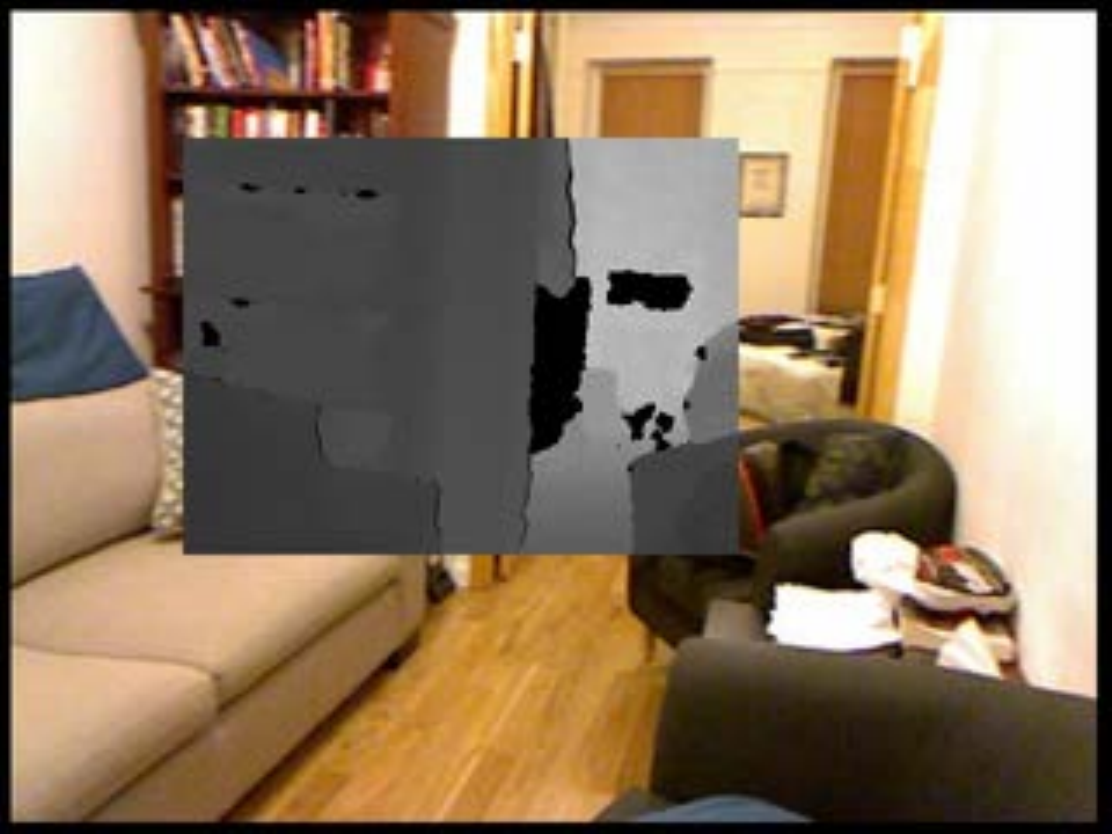} \\
  (a) Input image &
  (b) Depth &
  (c) Proposed method \\
  \includegraphics[clip,width=0.28\linewidth,bb=0 0 320 240]{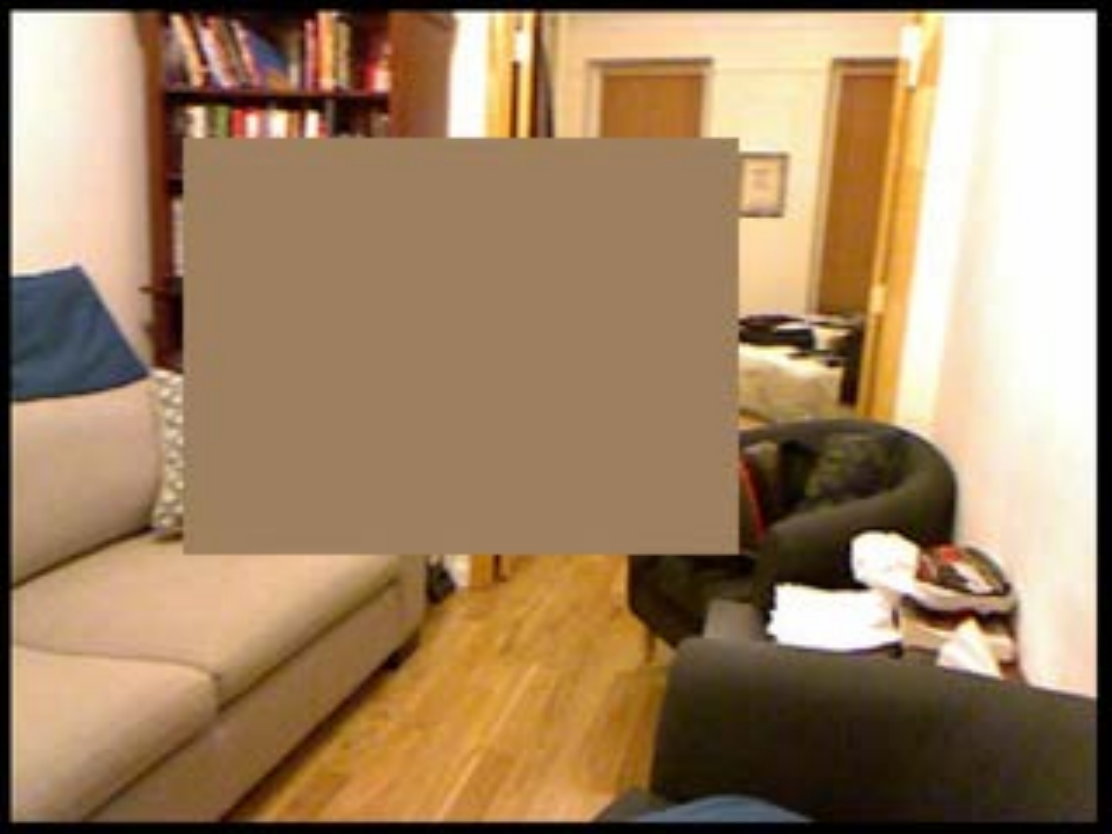} &
  \includegraphics[clip,width=0.28\linewidth,bb=0 0 320 240]{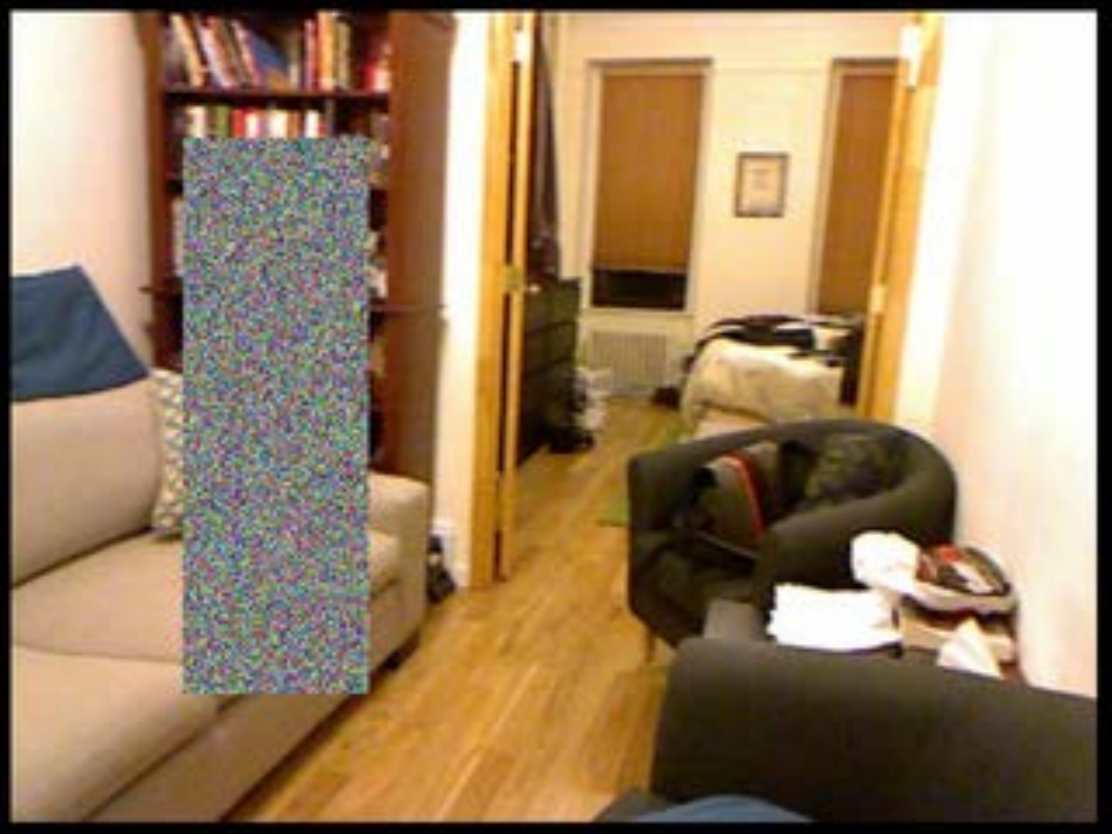} &
  \includegraphics[clip,width=0.28\linewidth,bb=0 0 320 240]{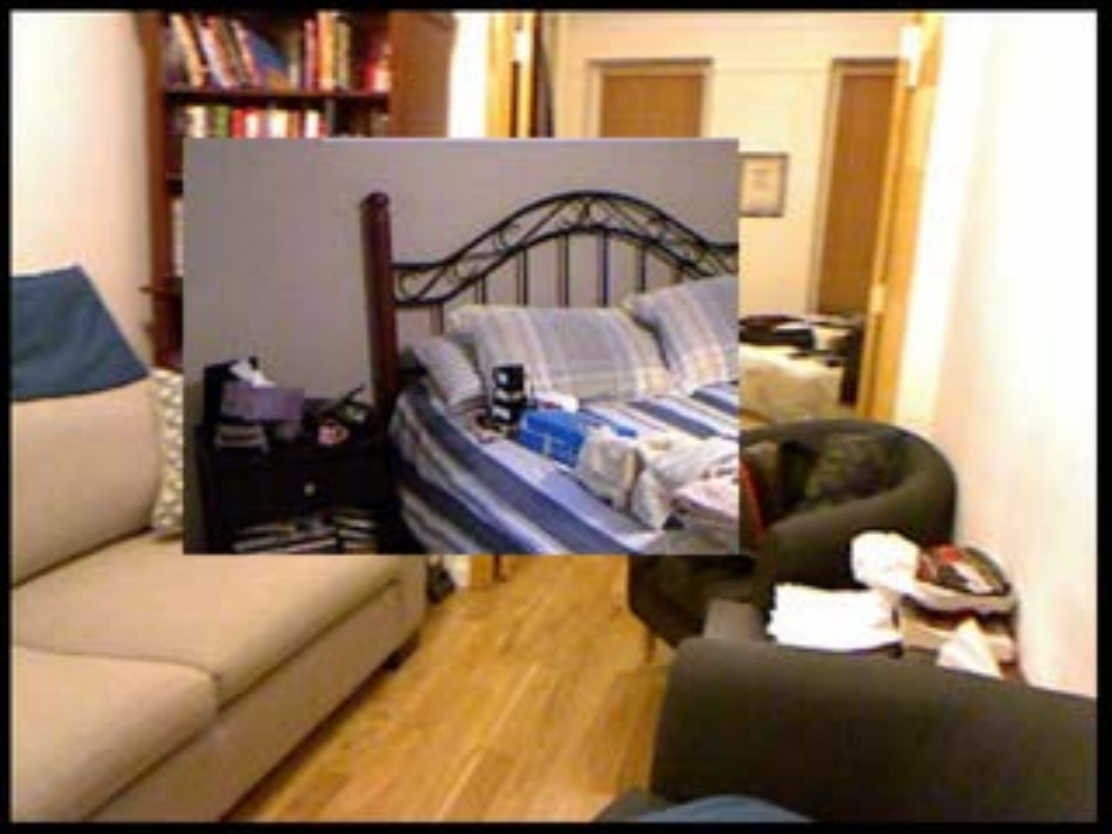} \\
  (d) CutOut &
  (e) RE &
  (f) CutMix
\end{tabular}
}
\caption{Examples of data augmentation}
\label{fig:DA}
\end{figure}
\section{Introduction}
Data augmentation is a practical way of improving the recognition performance without increasing the computational cost of inference.
Various data augmentation methods have been proposed in the field of computer vision \cite{devries2017improved, gong2020keepaugment, hendrycks2019benchmarking, yun2019cutmix, zhong2020random}.
These data augmentation methods have been studied for higher-order tasks (\eg, object recognition).
However, there has been little research on lower-order tasks, such as pixel-by-pixel transformations (\eg, monocular depth estimation).

The monocular depth estimation \cite{lee2019big, song2021monocular} estimates depth from a single-view RGB image.
This task requires the simultaneous acquisition of RGB images and depths.
Data augmentation is required because it is difficult to collect a large amount of data.
Image flipping, random cropping, and color and luminance transformation are often used for data augmentation.
However, few studies have examined data augmentation that changes the geometry because such augmentation reduces the estimation accuracy.

Yoo \etal \cite{yoo2020rethinking} proposed CutBlur, a data augmentation that adopts super-resolution.
CutBlur improves the estimation accuracy by pasting a partial region of a high-resolution image onto the same position of a low-resolution image.
The adoption of super-resolution generally has the problem of over-sharpening the estimated image.
CutBlur, however, has a regularization effect, which prevents excessive sharpening and reduces the estimation error.
Ghiasi \etal \cite{ghiasi2020simple} proposed a data augmentation for segmentation in which images are cut and pasted in units of the instance.
%The data augmentation does not adversely affect the feature extraction of the segmented object.

We propose CutDepth, which is a data augmentation that pastes the area cut from the teacher data (depth) at the same position on the input image (RGB image).
%CutDepth is a data augmentation method that mixes data from different domains( depth and RGB image ) into the input image (Figure \ref{fig:DA}).
By replacing a part of the input image with depth, 
the change in appearance is greater than that in conventional data augmentations.
Meanwhile, the change is small at the lower feature level because the edge positions of the depth and the RGB image are similar.
CutDepth regularizes the image by depth because depth information is given to the input image.
Therefore, the distance between the RGB image and the depth decreases in the latent space, and it becomes easier to estimate the depth.

The contributions of our work are as follows.
\begin{itemize}
    \item We propose a new data augmentation method that both improves visual diversity and suppresses excessive geometric changes in the scene.
    \item We show the quality of the data distribution after data augmentation in terms of diversity and affinity.
    \item We show that the depth estimation performance is improved subjectively and objectively for a real image using the proposed data augmentation method.
\end{itemize}

%%%%%%%%%%%%%%%%%%%%%
\section{Related work}
\subsection{Data augmentation}
Optical transformations and geometric transformations can be conducted for data augmentation \cite{shorten2019survey}.
The former transformations include changing luminance and colors whereas the latter transformations include image flipping, translation, affine transformation, and random clipping. 

There are methods of making changes optically and geometrically by replacing a partial area of the image with other information \cite{devries2017improved, yun2019cutmix, zhong2020random} (Figure \ref{fig:DA}).
CutOut \cite{devries2017improved} and Random Erasing (RE) \cite{zhong2020random} replace a portion of the image with the average value of the image or a random number.
CutMix \cite{yun2019cutmix} replaces a portion of an image with another image.

\subsection{Monocular depth estimation}
Monocular depth estimation \cite{lee2019big, song2021monocular} is the task of estimating the depth from a single-viewpoint image.
In monocular depth estimation, the input is an RGB image and the teacher data are of depth. 
BTS \cite{lee2019big} has a structure that implicitly estimates the normal in a decoder.
Monocular depth estimation has difficulty in estimating object contours.
However, the BTS structure improves the accuracy of the contour estimation.
Laplacian Depth \cite{song2021monocular} uses a Laplacian pyramid structure in a decoder.
The structure uses the residuals obtained from the input image for each resolution. 
It effectively estimates both local details and the global layout.

%
%%%%%%%%%%%%%%%
\section{Data augmentation for depth estimation}
\begin{figure}[t].
\centering
  \includegraphics[clip,width=1.0\linewidth, bb=0 0 980 580]{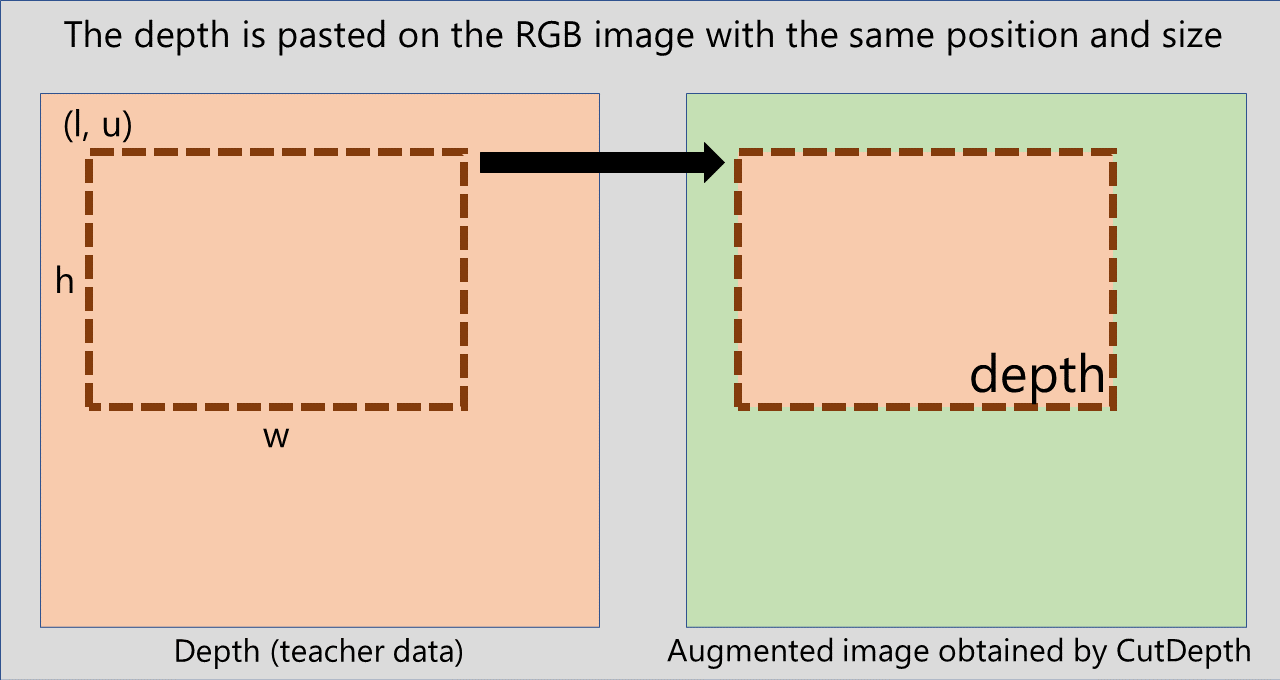}
\caption{Data augmentation using CutDepth}
\label{fig:proposedDA}
\end{figure}

We propose a data augmentation called CutDepth 
(Figure \ref{fig:proposedDA}).
Let $x_s \in R^{W \times H \times C_s}$ be the input (RGB) image and
$x_t \in R^{W \times H \times C_t}$ be the teacher (depth) data.
$W$ and $H$ are, respectively, the width and height of the image 
and $C_s$ and $C_t$ are, respectively, the numbers of channels in the input image and the teacher data.
$x'_s$ with data augmentation on $x_s$ is obtained as 
\begin{eqnarray}
x'_s = M * x_s + (1-M) * x_t.
\end{eqnarray}
If $C_s$ and $C_t$ are different, they are combined in the channel direction so that they are the same number in advance.
$M$ is a matrix ($M \in \{0,1\}$) that indicates the region where $x_s$ is replaced by $x_t$.
The position ($l$, $u$) and size ($w$, $h$) of the replacement region are obtained as 
\begin{align}
(l, u)&=&(a \times W , b \times H) \\
(w, h)&=&(\min((W - a \times W) \times c \times p , 1), \\
&& \min((H - b \times H) \times d \times p, 1)))
\end{align}
%where $a$, $b$, $c$, and $d$ are uniform random numbers $\mathcal{U}(0,1)$.
where $a$, $b$, $c$, and $d$ are $\mathcal{U}(0,1)$.
$p$ is a hyperparameter that determines the maximum values of $w$ and $h$ for $W$ and $H$, and it is set at a value of $(0,1]$.
%%%%%%%%%%%%%%
\begin{table*}[t]
  \label{tab:depth_est_result}
\centering
\caption{Comparison of the depth estimation performances when using different data augmentation methods. Lower Abs Rel, log10, RMSE and RMSE log indicate higher performance whereas higher d1, d2 and d3 indicate higher performance. The best performances are presented in bold text.}
\scalebox{0.7}{ 
\begin{tabular}{c|c|ccccccc||ccccccc}
  \multicolumn{2}{c}{}&\multicolumn{7}{c||}{BTS}&\multicolumn{7}{c}{Laplacian Depth} \\
  \hline
 Method & \textit{p} & Abs Rel $\downarrow$  & log10  $\downarrow$ & RMSE  $\downarrow$ & RMSE log  $\downarrow$ & d1 $\uparrow$ & d2 $\uparrow$ & d3 $\uparrow$ & Abs Rel $\downarrow$ & log10 $\downarrow$ & RMSE $\downarrow$ & RMSE log $\downarrow$ & d1 $\uparrow$ & d2 $\uparrow$ & d3 $\uparrow$ \\
  \hline
  Baseline & & 0.1122 &	0.048 &	0.406 & 0.145 & 0.878 & 0.979 & 0.995 & 0.11 & 0.047 & 0.39 & 0.139 & 0.884 & 0.983 & 0.996  \\
  \hline
  \multirow{4}{*}{CutOut} & 0.25 & 0.1122 & 0.048 & 0.405 & 0.144 & 0.878 & 0.98 & 0.996 & 0.106 & 0.046 & 0.384 & 0.136 & 0.891 & 0.984 & 0.996 \\
  & 0.50 & 0.1118 & 0.048 & 0.402 & 0.144 & 0.879 & 0.981 & 0.996   &  0.109 & 0.046 & 0.382 & 0.137 & 0.889 & 0.983 & \bf{0.997} \\
  & 0.75 & 0.1146 & 0.05 & 0.414 & 0.148 & 0.871 & 0.979 & 0.996 & 0.106 & 0.045 & 0.382 & 0.135 & 0.893 & \bf{0.985} & \bf{0.997} \\
  & 1.00 & 0.1194 & 0.051 & 0.427 & 0.152 & 0.864 & 0.977 & 0.996 &  0.11 & 0.047 & 0.394 & 0.14 & 0.884 & 0.984 & \bf{0.997}  \\
  \hline
    \multirow{2}{*}{Random} & 0.25 & 0.1106 & 0.048 & 0.4 & 0.143 & 0.88 & 0.981 & 0.996 & 0.109 & 0.046 & 0.384 & 0.137 & 0.89 & 0.982 & 0.996 \\
  \multirow{2}{*}{Erasing} & 0.50 & 0.1116 & 0.048 & 0.4 & 0.143 & 0.881 & 0.981 & 0.996  & 0.106 & 0.045 & 0.378 & 0.134 & 0.892 & \bf{0.985} & \bf{0.997} \\
  & 0.75 & 0.1132 & 0.049 & 0.415 & 0.147 & 0.871 & 0.979 & 0.996 & 0.106 & 0.045 & 0.379 & 0.134 & 0.893 & \bf{0.985} & \bf{0.997}\\
  & 1.00 & 0.1186 & 0.051 & 0.429 & 0.152 & 0.863 & 0.977 & 0.996 &
  0.111 & 0.047 & 0.394 & 0.14 & 0.884 & 0.983 & \bf{0.997} \\
  \hline
      \multirow{4}{*}{CutMix} & 0.25 & 0.1105 & 0.047 & 0.397 & 0.142 & 0.882 & 0.981 & 0.996 & 0.107 & 0.046 & 0.388 & 0.137 & 0.889 & 0.983 & 0.996 \\
  & 0.50 & 0.1132 & 0.049 & 0.406 & 0.146 & 0.874 & 0.979 & 0.996 &  0.107 & 0.046 & 0.386 & 0.136 & 0.891 & 0.983 & 0.996 \\
  & 0.75 & 0.1231 & 0.054 & 0.438 & 0.158 & 0.848 & 0.976 & 0.996 & 0.107 & 0.046 & 0.386 & 0.136 & 0.891 & 0.983 & 0.996 \\
  & 1.00 & 0.1851 & 0.086 & 0.674 & 0.241 & 0.659 & 0.918 & 0.982 & 0.11 & 0.047 & 0.391 & 0.139 & 0.886 & 0.982 & 0.996  \\
  \hline 
  \multirow{4}{*}{Proposed} & 0.25 & 0.1083 & 0.047 & 0.398 & 0.141 & 0.884 & 0.981 & 0.996 &  0.106 & 0.045 & 0.38 & 0.135 & 0.895 & 0.984 & 0.996 \\
  & 0.50 & 0.1077 & \bf{0.046} & \bf{0.391} & \bf{0.14} & 0.884 & \bf{0.982} & \bf{0.997} & \bf{0.104} & \bf{0.044} & \bf{0.375} & \bf{0.132} & \bf{0.899} & \bf{0.985} & \bf{0.997} \\
  & 0.75 & \bf{0.1074} & 0.047 & 0.392 & \bf{0.14} & \bf{0.885} & \bf{0.982} & 0.996 & 0.106 & 0.045 & 0.379 & 0.135 & 0.894 & 0.984 & \bf{0.997} \\
  & 1.00 & 0.1127 & 0.047 & 0.392 & 0.142 & 0.88 & 0.981 & 0.996 & \bf{0.104} & 0.045 & 0.376 & \bf{0.132} & 0.898 & \bf{0.985} & 0.996 
  \end{tabular}
  }   
\end{table*}

\begin{table}[t]
  \label{tab:depth_est_result_data_size}
\centering
\caption{Comparison of the depth estimation performances when using different numbers of data ($p$ = 0.75). Lower Abs Rel, log10, RMSE and RMSE log indicate higher performance whereas higher d1, d2 and d3 indicate higher performance.}
\scalebox{0.6}{ %ƒRƒR
\begin{tabular}{c|c|ccccccc}
%  \multicolumn{2}{c}{}&\multicolumn{7}{c}{BTS}\\
%  \hline
 Scale & Method & Abs Rel $\downarrow$ & log10  $\downarrow$ & RMSE  $\downarrow$ & RMSE log  $\downarrow$ & d1 $\uparrow$ & d2 $\uparrow$ & d3 $\uparrow$ \\
  \hline
\multirow{5}{*}{25\%} & Baseline & 0.1226 & \bf{0.052} & 0.428 & 0.154 & \bf{0.859} & 0.977 & 0.995 \\
& CutOut  &0.1242 & 0.053 & 0.432 & 0.156 & 0.854 & 0.976 & \bf{0.996}  \\
& RE & 0.1268 & 0.054 & 0.440& 0.158 & 0.848 & 0.976 & 0.995  \\
& CutMix & 0.1467 & 0.064 & 0.520 & 0.188 & 0.782 & 0.956 & 0.993 \\
& Proposed & \bf{0.1225} & \bf{0.052} & \bf{0.424} & \bf{0.153} & 0.858 & \bf{0.978} & 0.995  \\
\hline  
\multirow{5}{*}{50\%} & Baseline & 0.1174 & 0.050 & 0.414 & 0.150 & 0.867 & 0.978 & 0.995 \\
& CutOut  & 0.1168 & 0.050 & 0.418 & 0.150 & 0.867 & 0.979 & \bf{0.996}   \\
& RE & 0.1184 & 0.051 & 0.422 & 0.151 & 0.862 & 0.978 & \bf{0.996} \\
& CutMix & 0.1307 & 0.056 & 0.460 & 0.168 & 0.832 & 0.970 & 0.994  \\
& Proposed & \bf{0.1155} & \bf{0.049} & \bf{0.411} & \bf{0.148} & \bf{0.870} & \bf{0.981} & \bf{0.996}  \\
\hline  
\multirow{5}{*}{75\%} & Baseline & 0.1154 & 0.049 & 0.410 &  0.147 & 0.871 & 0.979 & 0.996  \\
& CutOut  & 0.1148 & 0.050 & 0.413 & 0.147 & 0.870 & 0.980 & \bf{0.997}
 \\
& RE & 0.1179 & 0.051 & 0.424 & 0.151 & 0.863 & 0.977 & 0.996 \\
& CutMix & 0.1353 & 0.058 & 0.465 &  0.172 & 0.826 & 0.967 & 0.993 \\
& Proposed & \bf{0.1142} & \bf{0.048} & \bf{0.401} & \bf{0.144} & \bf{0.876} & \bf{0.981} & 0.996 \\
%  \multirow{4}{*}{Proposed} & 0.25 & 0.1083 & 0.047 & 0.398 & 0.141 & 0.884 & 0.981 & 0.996 &  0.106 & 0.045 & 0.38 & 0.135 & 0.895 & 0.984 & 0.996 \\
  \end{tabular}
  }
\end{table}
\section{Experimental results}
\subsection{Experimental setting}
\label{sec:exp_env}
We use BTS \cite{lee2019big} and Laplacian Depth \cite{song2021monocular} to evaluate the performance of the depth estimation.
The optimizers are Adam and AdamW  respectively.
The learning rate is $10^{-4}$ for both, and the scheduling is performed with a polynomial decay (power of 0.9 and 0.5, respectively).
The encoders are DenseNet161 \cite{huang2017densely} and ResNext-101 \cite{xie2017aggregated}, pre-trained on ImageNet.
The baseline data augmentation method uses horizontal flipping, color transformations, and image rotation randomly.

We use the NYU Depth V2 Dataset \cite{silberman2012indoor}.
%Since there are various objects indoors and the variation of the test scene is large, the generalization performance is important.
The image size used for training is 416 $\times$ 544 and that used for testing is 480 $\times$ 640, as in BTS \cite{lee2019big}.
As in BTS, 24,231 RGB image/depth pairs are used for training and 654 images are used for evaluation.
%24,231 images are used for training and 654 images are used for evaluation.

\subsection{Depth estimation results}
In addition to the baseline method, we use
CutOut \cite{devries2017improved}, RE \cite{zhong2020random}, and CutMix \cite{yun2019cutmix}.
Table \ref{tab:depth_est_result} gives the evaluation results.
The proposed method outperforms the conventional methods on all metrics.
The depth estimation performance of the proposed method tended to be higher when $p$ was 0.5 or 0.75.
The performances of the conventional methods decreased as $p$ increased.
The performance of the proposed method did not degrade with increasing $p$ owing to the small change in the lower-order feature levels.
%In the proposed method, the edges of the input image and the depth image are similar.

We next compare the depth estimation performances when the number of data are 25 \%, 50 \%, and 75 \% of the original number and $p$ is set at 0.75.
We randomly sample the data from the original dataset.
Table \ref{tab:depth_est_result_data_size} gives the evaluation results for different sizes of data.
Even if the data size was small, the proposed method outperformed the other methods on various measures.

\begin{figure*}[t]
\centering
\scalebox{0.9}{ %ƒRƒR
\begin{tabular}{ccccccc}
\multicolumn{7}{c}{BTS} \\
      \includegraphics[clip,width=0.12\linewidth,bb=0 0 320 240]{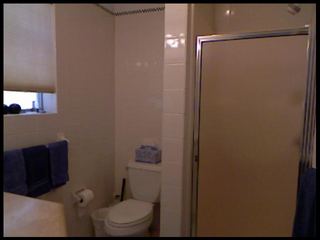} &
  \includegraphics[clip,width=0.12\linewidth,bb=0 0 320 240]{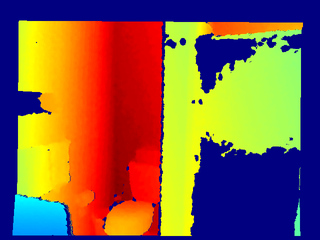} &
  \includegraphics[clip,width=0.12\linewidth,bb=0 0 320 240]{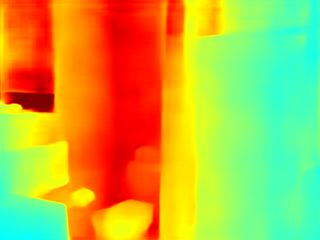} &
  \includegraphics[clip,width=0.12\linewidth,bb=0 0 320 240]{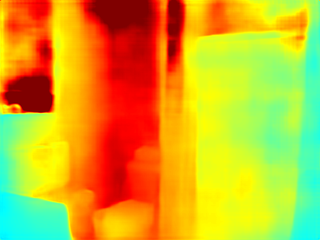} &
  \includegraphics[clip,width=0.12\linewidth,bb=0 0 320 240]{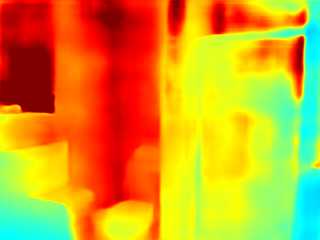} &
  \includegraphics[clip,width=0.12\linewidth,bb=0 0 320 240]{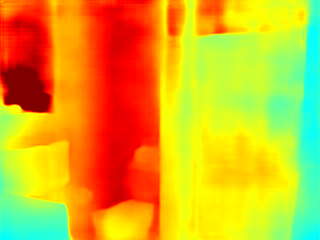} &
  \includegraphics[clip,width=0.12\linewidth,bb=0 0 320 240]{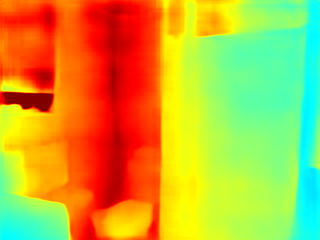}   
    \\
      \includegraphics[clip,width=0.12\linewidth,bb=0 0 320 240]{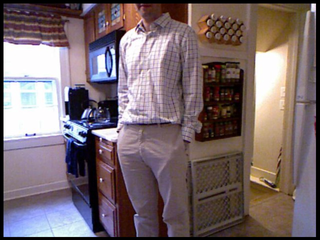} &
  \includegraphics[clip,width=0.12\linewidth,bb=0 0 320 240]{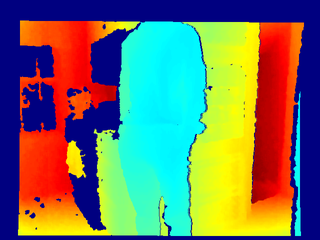} &
  \includegraphics[clip,width=0.12\linewidth,bb=0 0 320 240]{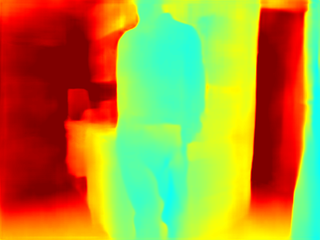} &
  \includegraphics[clip,width=0.12\linewidth,bb=0 0 320 240]{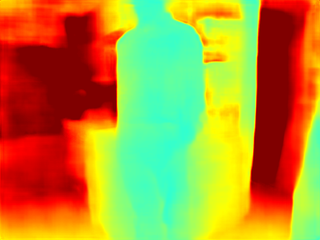} &
  \includegraphics[clip,width=0.12\linewidth,bb=0 0 320 240]{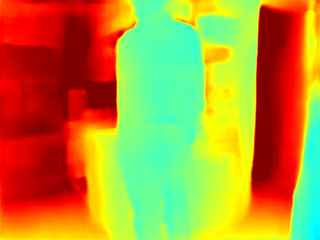} &
  \includegraphics[clip,width=0.12\linewidth,bb=0 0 320 240]{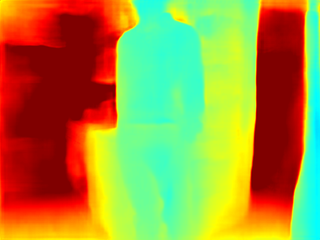} &
  \includegraphics[clip,width=0.12\linewidth,bb=0 0 320 240]{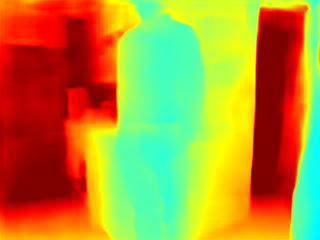} \\
     \includegraphics[clip,width=0.12\linewidth,bb=0 0 320 240]{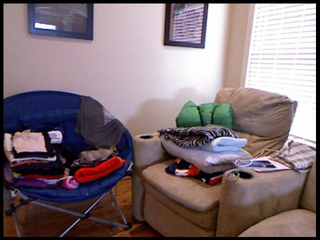} &
  \includegraphics[clip,width=0.12\linewidth,bb=0 0 320 240]{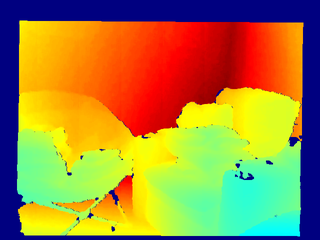} &
  \includegraphics[clip,width=0.12\linewidth,bb=0 0 320 240]{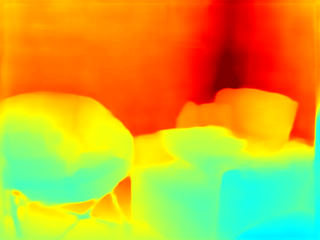} &
  \includegraphics[clip,width=0.12\linewidth,bb=0 0 320 240]{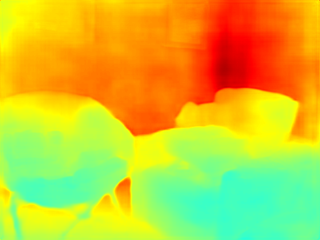} &
  \includegraphics[clip,width=0.12\linewidth,bb=0 0 320 240]{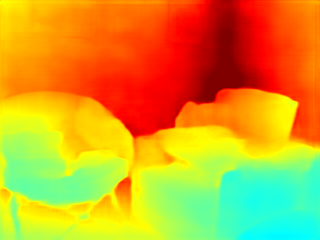} &
  \includegraphics[clip,width=0.12\linewidth,bb=0 0 320 240]{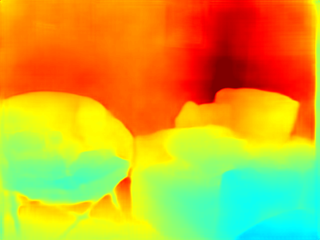} &
  \includegraphics[clip,width=0.12\linewidth,bb=0 0 320 240]{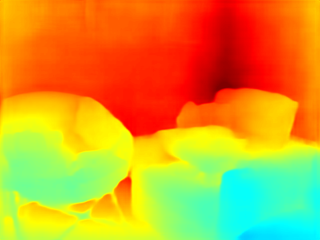} \\
      \includegraphics[clip,width=0.12\linewidth,bb=0 0 320 240]{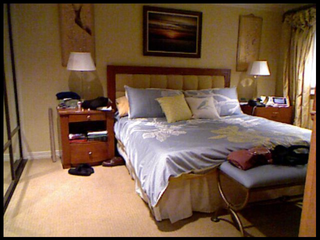} &
  \includegraphics[clip,width=0.12\linewidth,bb=0 0 320 240]{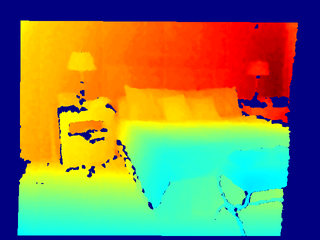} &
  \includegraphics[clip,width=0.12\linewidth,bb=0 0 320 240]{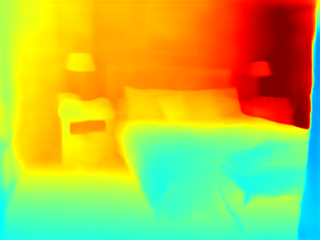} &
  \includegraphics[clip,width=0.12\linewidth,bb=0 0 320 240]{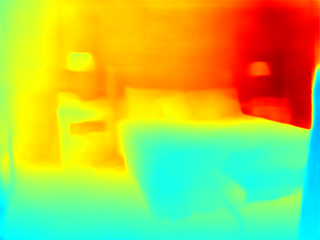} &
  \includegraphics[clip,width=0.12\linewidth,bb=0 0 320 240]{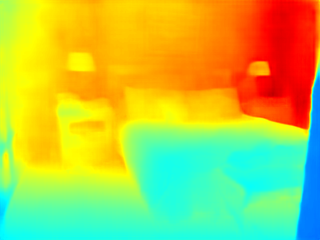} &
  \includegraphics[clip,width=0.12\linewidth,bb=0 0 320 240]{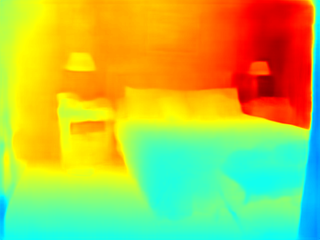} &
  \includegraphics[clip,width=0.12\linewidth,bb=0 0 320 240]{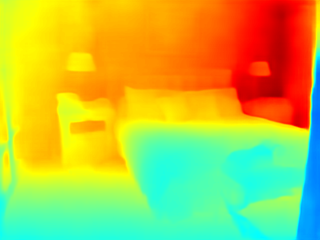} \\
    \includegraphics[clip,width=0.12\linewidth,bb=0 0 320 240]{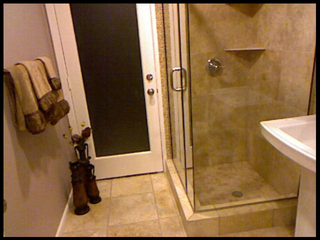} &
  \includegraphics[clip,width=0.12\linewidth,bb=0 0 320 240]{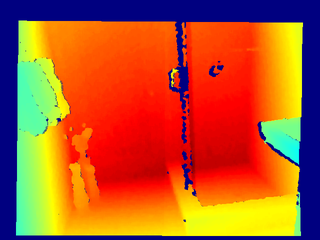} &
  \includegraphics[clip,width=0.12\linewidth,bb=0 0 320 240]{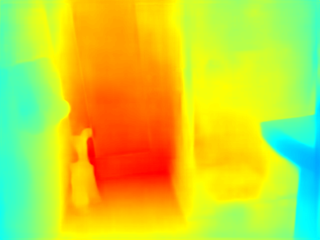} &
  \includegraphics[clip,width=0.12\linewidth,bb=0 0 320 240]{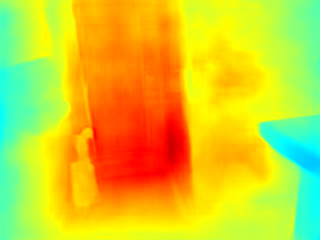} &
  \includegraphics[clip,width=0.12\linewidth,bb=0 0 320 240]{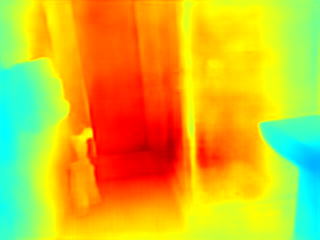} &
  \includegraphics[clip,width=0.12\linewidth,bb=0 0 320 240]{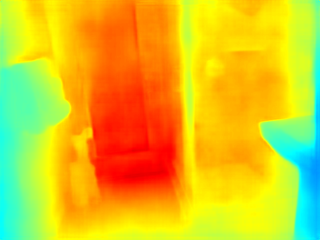} &
  \includegraphics[clip,width=0.12\linewidth,bb=0 0 320 240]{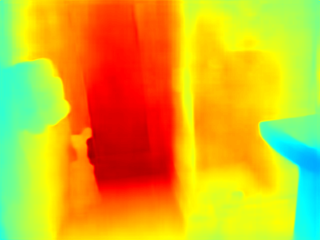} \\
\\
\multicolumn{7}{c}{laplacian depth} \\
 \includegraphics[clip,width=0.12\linewidth,bb=0 0 320 240]{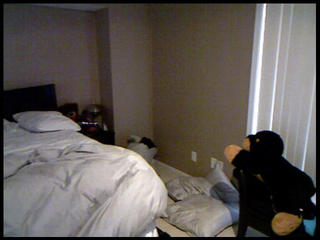} &
  \includegraphics[clip,width=0.12\linewidth,bb=0 0 320 240]{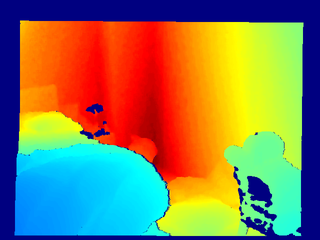} &
  \includegraphics[clip,width=0.12\linewidth,bb=0 0 320 240]{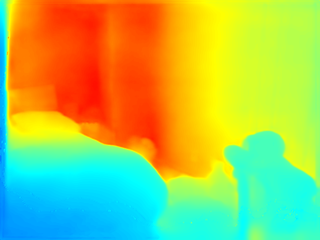} &
  \includegraphics[clip,width=0.12\linewidth,bb=0 0 320 240]{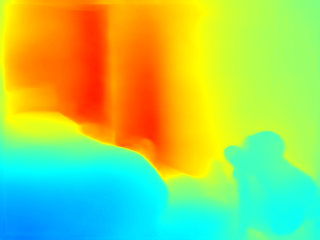} &
  \includegraphics[clip,width=0.12\linewidth,bb=0 0 320 240]{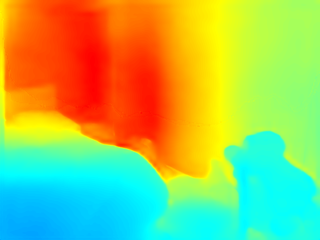} &
  \includegraphics[clip,width=0.12\linewidth,bb=0 0 320 240]{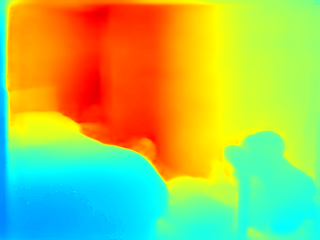} &
  \includegraphics[clip,width=0.12\linewidth,bb=0 0 320 240]{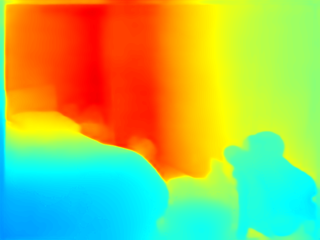} \\
 \includegraphics[clip,width=0.12\linewidth,bb=0 0 320 240]{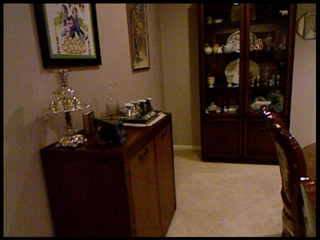} &
  \includegraphics[clip,width=0.12\linewidth,bb=0 0 320 240]{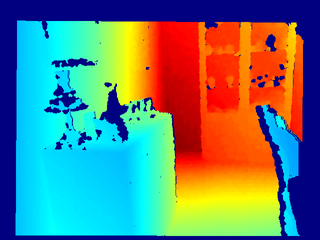} &
  \includegraphics[clip,width=0.12\linewidth,bb=0 0 320 240]{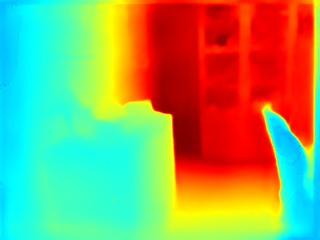} &
  \includegraphics[clip,width=0.12\linewidth,bb=0 0 320 240]{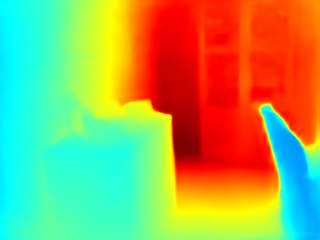} &
  \includegraphics[clip,width=0.12\linewidth,bb=0 0 320 240]{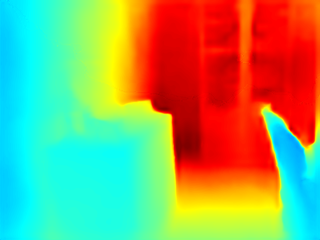} &
  \includegraphics[clip,width=0.12\linewidth,bb=0 0 320 240]{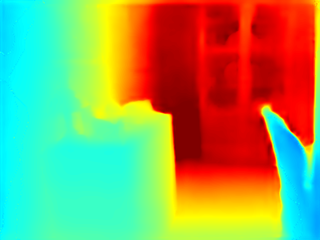} &
  \includegraphics[clip,width=0.12\linewidth,bb=0 0 320 240]{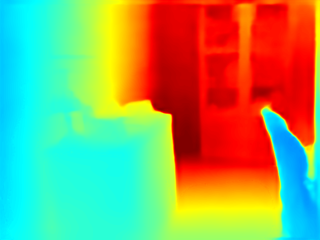} \\
     \includegraphics[clip,width=0.12\linewidth,bb=0 0 320 240]{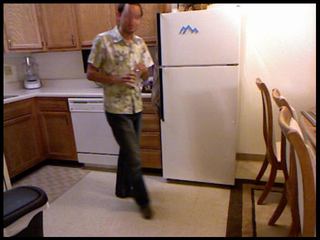} &
  \includegraphics[clip,width=0.12\linewidth,bb=0 0 320 240]{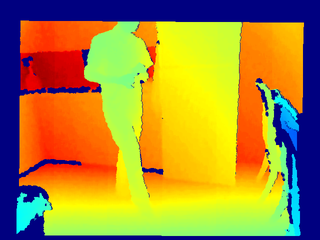} &
  \includegraphics[clip,width=0.12\linewidth,bb=0 0 320 240]{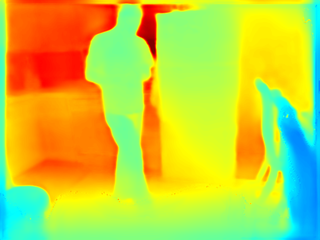} &
  \includegraphics[clip,width=0.12\linewidth,bb=0 0 320 240]{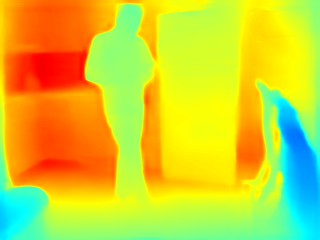} &
  \includegraphics[clip,width=0.12\linewidth,bb=0 0 320 240]{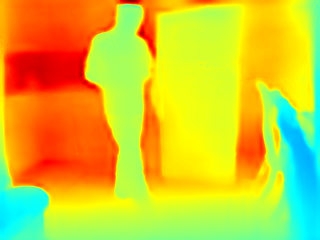} &
  \includegraphics[clip,width=0.12\linewidth,bb=0 0 320 240]{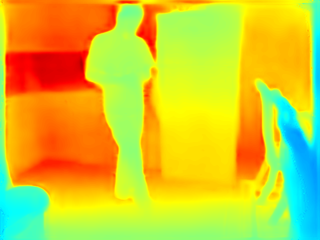} &
  \includegraphics[clip,width=0.12\linewidth,bb=0 0 320 240]{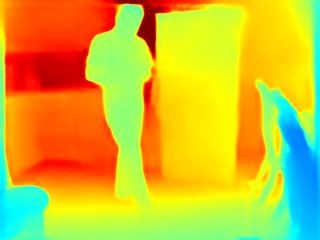} \\

  \includegraphics[clip,width=0.12\linewidth,bb=0 0 320 240]{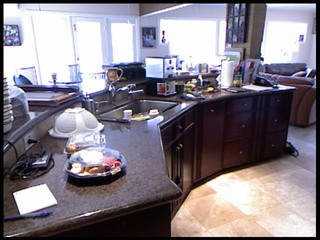} &
  \includegraphics[clip,width=0.12\linewidth,bb=0 0 320 240]{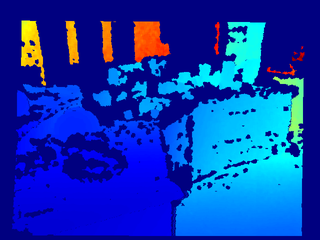} &
  \includegraphics[clip,width=0.12\linewidth,bb=0 0 320 240]{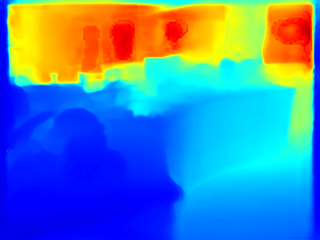} &
  \includegraphics[clip,width=0.12\linewidth,bb=0 0 320 240]{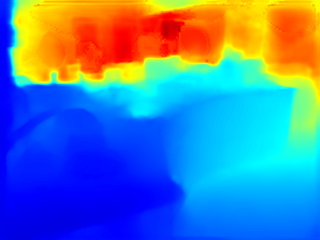} &
  \includegraphics[clip,width=0.12\linewidth,bb=0 0 320 240]{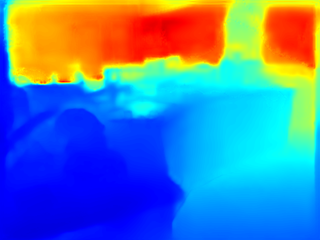} &
  \includegraphics[clip,width=0.12\linewidth,bb=0 0 320 240]{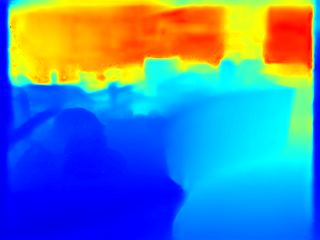} &
  \includegraphics[clip,width=0.12\linewidth,bb=0 0 320 240]{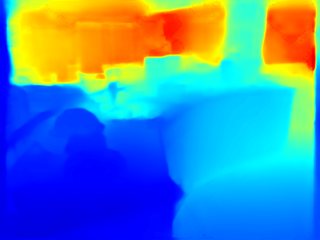} \\
   \includegraphics[clip,width=0.12\linewidth,bb=0 0 320 240]{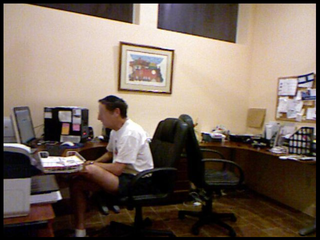} &
  \includegraphics[clip,width=0.12\linewidth,bb=0 0 320 240]{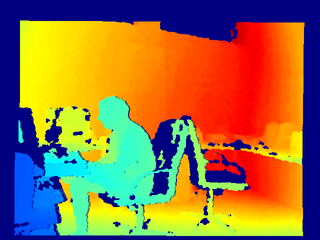} &
  \includegraphics[clip,width=0.12\linewidth,bb=0 0 320 240]{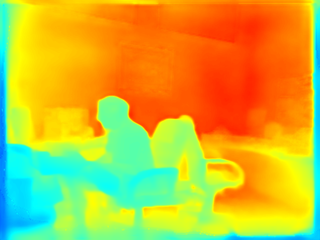} &
  \includegraphics[clip,width=0.12\linewidth,bb=0 0 320 240]{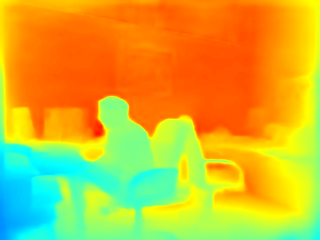} &
  \includegraphics[clip,width=0.12\linewidth,bb=0 0 320 240]{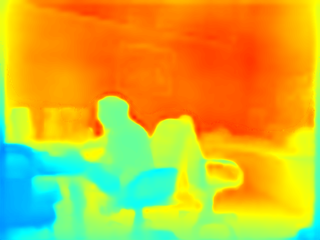} &
  \includegraphics[clip,width=0.12\linewidth,bb=0 0 320 240]{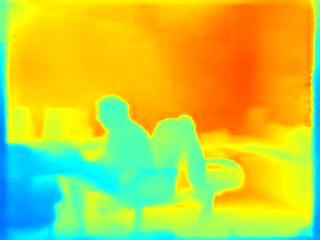} &
  \includegraphics[clip,width=0.12\linewidth,bb=0 0 320 240]{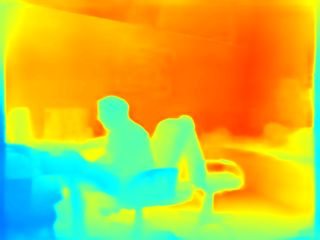} \\
   (a) RGB image & (b) Ground Truth & (c) Baseline & (d) CutMix & (e) CutOut  & (f) RE & (g) Proposed
  \end{tabular}
  }
\caption{Depth estimation results obtained using different data augmentation methods}
\label{fig:result_img}
\end{figure*}

Figure \ref{fig:result_img} shows example results of depth estimation.
Near distances are represented in blue and far distances in red.
The proposed method outperformed the conventional methods in terms of the accuracy of the estimation for far distances and object contours.
The conventional training model readily overfit some data because there were few data at a distance.
However, the proposed method had better accuracy of the data at a distance because overfitting was reduced by regularization.
\subsection{Evaluation of the effect of regularization}
The distance between the RGB image and depth in the latent space becomes small if the image is regularized by depth.
Therefore, to verify the regularization effect, we compare the distances in the latent space, which are the output of the BTS encoder, when the RGB image and depth are input to the BTS model.
%
%We calculate the distance using the feature map output by BTS encoder.
The root-mean-square error (RMSE), mean absolute error (MAE), and cosine distance are used as distance measures.
Table \ref{tab:bottleneck_dis} gives the comparison results.
In terms of the RMSE and MAE, the distances of the proposed method and CutMix are comparable.
However, the cosine distance is small for the proposed method.
It is difficult to see the difference between the RMSE and MAE because of the small scale of the feature map.
However, the difference becomes clear for the cosine distance where the scale is normalized.
\subsection{Evaluation of the properties of data augmentation}
\begin{figure}[t]
\centering
  \includegraphics[clip, width=1.0\linewidth,bb=0 0 556 756]{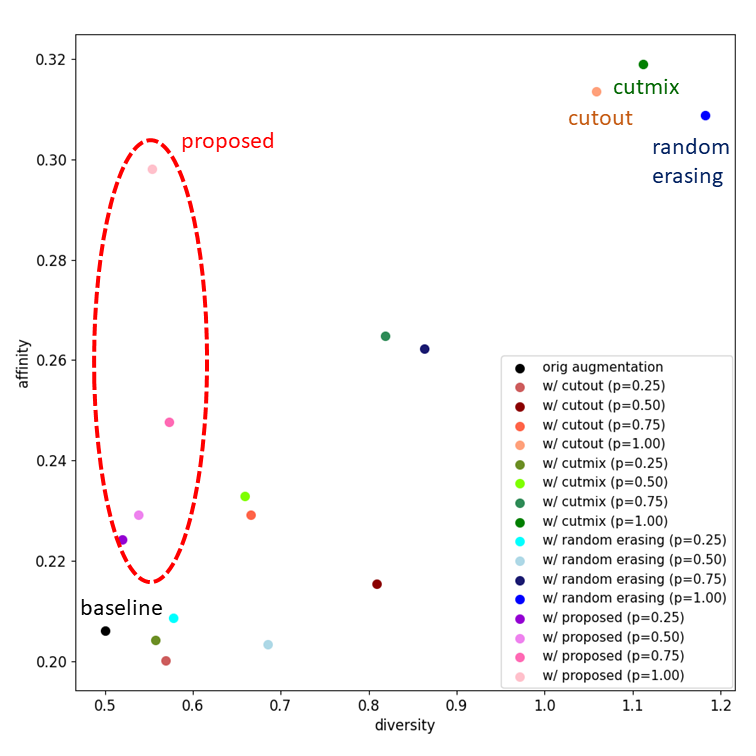}
\caption{Comparison of diversity and affinity between different data augmentation methods}
\label{fig:affinity}
\end{figure}

We examine the properties of data augmentation.
Gontijo-Lopes \etal \cite{gontijo2020affinity} proposed the measurement of the properties of data augmentation in terms of diversity and affinity.
A larger value of diversity corresponds to a greater spread of the data distribution due to data augmentation.
A larger value of affinity corresponds to a smaller deviation from the original data distribution.
Figure \ref{fig:affinity} plots the diversity and affinity of each method.
Pink circles show the results of the proposed method and are encircled by a red dotted ellipse for clarity.
Both the diversity and affinity of the proposed method are greater than those of the baseline method, and the distribution is thus more spread out and the deviation is smaller than for the original data.
The proposed method has lower diversity than the various conventional methods, and the effect of suppressing excessive changes in edge features is thus confirmed.

\begin{table}[t]
\centering
\caption{Comparison of the distances between the RGB image and depth in the latent space}
\scalebox{1.0}{ %ƒRƒR
    \begin{tabular}{c|c|ccc}
    &  \textit{p} & RMSE $\downarrow$ & MAE $\downarrow$ & Cosine $\uparrow$ \\
    \hline
    Baseline & & 1.094 & 0.49 & 0.24 \\
    \hline
    \multirow{4}{*}{CutOut} & 0.25 & 1.12 & 0.50 & 0.21 \\
    & 0.50 & 1.16 & 0.52 & 0.17 \\
    & 0.75 & 1.20 & 0.52 & 0.17 \\
    & 1.00 & 1.39 & 0.61 & 0.15 \\
    \hline    
    \multirow{2}{*}{Random} & 0.25 & 1.05 & 0.48 & 0.22 \\
    \multirow{2}{*}{Erasing} & 0.50 & 1.09 & 0.49 & 0.20  \\
    & 0.75 & 1.13 & 0.50 & 0.17 \\
    & 1.00 & 1.17 & 0.52 & 0.17 \\
    \hline    
    \multirow{4}{*}{CutMix} & 0.25 & 1.03 & 0.47 & 0.28 \\
    & 0.50 & \bf{0.92} & \bf{0.41} & 0.22 \\
    & 0.75 & 0.95 & 0.43 & 0.20 \\
    & 1.00 & 1.35 & 0.50 & 0.12 \\
    \hline    
    \multirow{4}{*}{Proposed} & 0.25 & \bf{0.92} & 0.42 & \bf{0.37} \\
    & 0.50 & 1.06 & 0.48 & \bf{0.37} \\
    & 0.75 & 0.96 & 0.44 & 0.35 \\
    & 1.00 & 1.07 & 0.48 & 0.33 \\
    \end{tabular}
    }
    \label{tab:bottleneck_dis}

\end{table}

%
%%%%%%%%%%%%%%%%%
\section{Conclusion}
We proposed a data augmentation method, called CutDepth, for depth estimation.
CutDepth is a method of pasting part of the depth to an input image, 
which increases the variation of the input image.
We confirmed that the estimation accuracy of the proposed method is better than that of conventional methods.
%In particular, it was confirmed that the accuracy of far distance and object contours with little training data was improved.
In the proposed method, the edge features are similar before and after the data augmentation.
We therefore found that the proposed method does not expand the data distribution excessively compared with the conventional methods.
In future work, we will test the effectiveness of the proposed method in tasks other than depth estimation.
%-------------------------------------------------------------------------
% \eg \etal
% citation‚Í”Žš‚Ì‡”Ô
%\begin{figure*}
%\begin{center}
%\fbox{\rule{0pt}{2in} \rule{.9\linewidth}{0pt}}
%\end{center}
%   \caption{Example of a short caption, which should be centered.}
%\label{fig:short}
%\end{figure*}

%------------------------------------------------------------------------

%-------------------------------------------------------------------------

{\small
\bibliographystyle{ieee_fullname}
\bibliography{main}
}

\end{document}